# On the Detection of Concept Changes in Time-Varying Data Stream by Testing Exchangeability

**Shen-Shyang Ho**
Dept. of Computer Science
George Mason University
Fairfax, VA 22030

**Harry Wechsler**
Dept. of Computer Science
George Mason University
Fairfax, VA 22030

## Abstract

A martingale framework for concept change detection based on testing data exchangeability was recently proposed (Ho, 2005). In this paper, we describe the proposed change-detection test based on the Doob's Maximal Inequality and show that it is an approximation of the sequential probability ratio test (SPRT). The relationship between the threshold value used in the proposed test and its size and power is deduced from the approximation. The mean delay time before a change is detected is estimated using the average sample number of a SPRT. The performance of the test using various threshold values is examined on five different data stream scenarios simulated using two synthetic data sets. Finally, experimental results show that the test is effective in detecting changes in time-varying data streams simulated using three benchmark data sets.

## 1 Introduction

A recent challenge in online mining of data streams is the detection of changes in the data-generating process. In the data stream, the target concept to be learned may change over time, known commonly as concept drift. These changes can be either gradual or abrupt. By locating these changes, one gains the knowledge of the instance when concept change occurs. Moreover, the data model can be adapted based on these located changes. These changes in concept in a data stream have an important characteristic: the violation of data exchangeability condition in the data stream.

The problem of detecting changes in sequential data was first studied by statisticians and mathematicians. In the online setting, data are observed one by one from a source. The disruption of "stochastic homogeneity" of the data might signal a change in the data generating process which would require decision making to avoid possible losses. This problem is generally known as "change-point detection". Change detection methods first appeared based on Wald's sequential probability ratio test (SPRT) (Wald, 1947). Later, Page introduced the cumulative sum method (Page, 1957) and Girshik and Rubin (Girshik & Rubin, 1952) also developed a test to detect changes. These methods, however, make strong parametric assumptions about the data distribution and may not be suitable for real data. Recently, Kifer et al. (2004) proposed a non-parametric change-detection method with statistical guarantees on the reliability of detected changes, but the use of their discrepancy measure on high dimensional data streams is still under investigation.

The contribution of this paper is the justification that the change-detection test based on the Doob's Maximal Inequality is an approximation of the sequential probability ratio test (SPRT), which is then used to (i) deduce the relation between the threshold value used in the change-detection test and its size and power, and (ii) estimate the mean delay time before a change is detected.

In Section 2, we review the concept of martingale and exchangeability and then, in Section 3, we describe the change detection method using the martingale. In Section 4, we show that the martingale test is an approximation of the SPRT. We examine the change detection method on time-varying data streams simulated using two synthetic data sets and three benchmark data sets in Section 5.

## 2 Martingale and Exchangeability

Let $\{Z_i : 1 \leq i < \infty\}$ be a sequence of random variables. A finite sequence of random variables $Z_1, \cdots, Z_n$ is *exchangeable* if the joint distribu-tion $p(Z_1, \cdots, Z_n)$ is invariant under any permutation of the indices of the random variables. A *martingale* is a sequence of random variables $\{M_i : 0 \leq i < \infty\}$ such that $M_n$ is a measurable function of $Z_1, \cdots, Z_n$ for all $n = 0, 1, \cdots$ (in particular, $M_0$ is a constant value) and the conditional expectation of $M_{n+1}$ given $M_0, \cdots, M_n$ is equal to $M_n$, i.e.

$$E(M_{n+1}|M_1, \cdots, M_n) = M_n \quad (1)$$

Vovk et al. (2003) introduced the idea of testing exchangeability online using the martingale. After observing a new data point, a learner outputs a positive martingale value reflecting the strength of evidence found against the null hypothesis of data exchangeability. Consider a set of labeled examples $Z = \{z_1, \cdots, z_{n-1}\} = \{(x_1, y_1), \cdots, (x_{n-1}, y_{n-1})\}$ where $x_i$ is an object and $y_i \in \{-1, 1\}$, its corresponding label, for $i = 1, 2, \cdots, n-1$. Assuming that a new labeled example, $z_n$, is observed, testing exchangeability for the sequence of examples $z_1, z_2, \cdots, z_n$ consists of two main steps (Vovk et al., 2003):

**A. Extract a p-value $p_n$ for the set $Z \cup \{z_n\}$ from the strangeness measure deduced from a classifier**

The randomized p-value of the set $Z \cup \{z_n\}$ is define as

$$V(Z \cup \{z_n\}, \theta_n) = \frac{\#\{i : \alpha_i > \alpha_n\} + \theta_n \#\{i : \alpha_i = \alpha_n\}}{n} \quad (2)$$

where $\alpha_i$ is the strangeness measure for $z_i$, $i = 1, 2, \cdots, n$ and $\theta_n$ is randomly chosen from $[0, 1]$. The strangeness measure is a way of scoring how a data point is different from the rest. Each data point $z_i$ is assigned a strangeness value $\alpha_i$ based on the classifier used (e.g. support vector machine (SVM), nearest neighbor rule, and decision tree). In our work, the SVM is used to compute the strangeness measure, which can be either the Lagrange multipliers or the distances from the hyperplane for the examples in $Z \cup \{z_n\}$.

The p-values $p_1, p_2, \cdots$ output by the randomized p-value function $V$ are distributed uniformly in $[0, 1]$, provided that the input examples $z_1, z_2, \cdots$ are generated by an exchangeable probability distribution in the input space (Vovk et al., 2003). This property of output p-values no longer holds when the exchangeability condition is not satisfied (see Section 3).

**B. Construct the randomized power martingale**

A family of martingales, indexed by $\epsilon \in [0, 1]$, and referred to as the *randomized power martingale*, is defined as

$$M_n^{(\epsilon)} = \prod_{i=1}^n \left(\epsilon p_i^{\epsilon-1}\right) \quad (3)$$

where the $p_i$s are the p-values output by the randomized p-value function $V$, with the initial martingale $M_0^{(\epsilon)} = 1$. We note that $M_n^{(\epsilon)} = \epsilon p_n^{\epsilon-1} M_{n-1}^{(\epsilon)}$. Hence, it is not necessary to stored the previous p-values. In our experiments, we use $\epsilon = 0.92$, which is within the desirable range where the martingale value is more sensitive to a violation of the exchangeability condition (Vovk et al., 2003). The role of $\epsilon$, with respect to the characteristics of a likelihood ratio, will be discussed in Section 4.

## 3 Change Detection using Martingale

Intuitively, we assume that a sequence of data points with a concept change consists of two concatenated data segments, $S_1$ and $S_2$, such that the concepts of $S_1$ and $S_2$ are $C_1$ and $C_2$ respectively and $C_1 \neq C_2$. Switching a data point $z_i$ from $S_2$ to a position in $S_1$ will make the data point stands out in $S_1$. The exchangeability condition is, therefore, violated. Exchangeability is a necessary condition for a conceptually stable data stream. The absence of exchangeability would suggest concept changes.

When a concept change occurs, the p-values output from the randomized p-value function (2) become skewed and the p-value distribution is no longer uniform. By the Kolmogorov-Smirnov Test [1,2], the p-values are shown not to be distributed uniformly after the concept changes. The null hypothesis "the p-values output by (2) are uniformly distributed" is rejected at significance level $\alpha = 0.05$, after sufficient number of data points (about $100 - 200$) are observed (see the example in Figure 1). The skewed p-value distribution plays an important role in the martingale test for change detection as small p-values inflate the martingale values. We note that an immediate detection of a true change is practically impossible.

For the martingale test, one decides whether a change occurs based on whether there is a violation of exchangeability condition which is, in turn, based on the martingale value. Consider the simple null hypothesis $H_0$ : "no concept change in the data stream" against the alternative $H_1$ : "concept change occurs in the data

---
[1]Kifer et al. (2004) proposed using Kolmogorov-Smirnov Test (KS-Test) for detecting changes using two sliding windows and a discrepancy measure which was tested only on 1D data stream.

[2]The readers should not confuse the p-values from the KS-Test in Figure 1 with the p-values from the randomized p-value function.

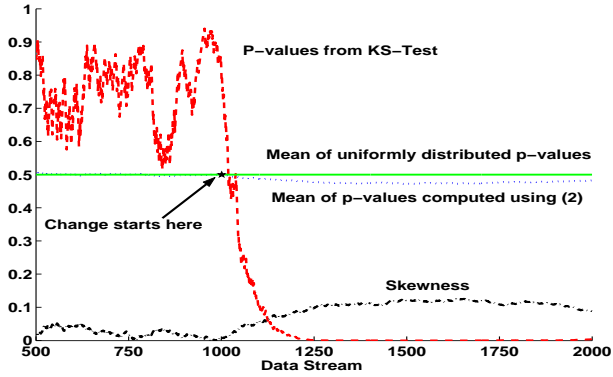

Table 1: The 10-dimensional data points simulated using normally distributed clusters data generator (see Section 5.1.2) are observed one by one from the $1st$ to the $2000th$ data point with concept change starting at the $1001th$ data point. (Y-axis represents values of the p-values, the skewness and the significance level.)

stream". The test continues to operate as long as

$$0 < M_n^{(\epsilon)} < \lambda \tag{4}$$

where $\lambda$ is a positive number. One rejects the null hypothesis when $M_n^{(\epsilon)} \geq \lambda$.

Assuming that $\{M_k : 0 \leq k < \infty\}$ is a nonnegative martingale, the Doob's Maximal Inequality (Steele, 2001) states that for any $\lambda > 0$ and $0 \leq n < \infty$,

$$\lambda P\left(\max_{k \leq n} M_k \geq \lambda\right) \leq E(M_n) \tag{5}$$

Hence, if $E(M_n) = E(M_1) = 1$, then

$$P\left(\max_{k \leq n} M_k \geq \lambda\right) \leq \frac{1}{\lambda} \tag{6}$$

This inequality means that it is unlikely for any $M_k$ to have a high value. One rejects the null hypothesis when the martingale value is greater than $\lambda$. But there is a risk of announcing a change detected when there is no change. The amount of risk one is willing to take will determine what $\lambda$ value to use.

## 4 Exchangeability Martingale Test as an Approximation to the Sequential Probability Ratio Test

Let $f(z, H_j)$ be the probability of observed point(s) $z$ given the hypothesis, $H_j$. For a fixed sample $Z = \{z_1, z_2, \cdots, z_n\}$, the most powerful test (smallest $\beta$ where $\beta$ is the probability of the type II error) depends on the likelihood ratio $l_n$, where

$$l_n = \prod_{i=1}^{n} \frac{f(z_i, H_1)}{f(z_i, H_0)} = \frac{f(Z, H_1)}{f(Z, H_0)} \tag{7}$$

and the test decides for or against the null hypothesis, $H_0$ according as $l_n$ is less than or greater than a chosen constant of a desirable size $\alpha$ and test power $(1 - \beta)$.

The sequential probability ratio test (SPRT) introduced by Wald (1947) for testing the null hypothesis $H_0 : \theta = \theta_0$ against the alternative hypothesis $H_1 : \theta = \theta_1$ based on observations $z_1, z_2, \cdots$ is analogous to the above likelihood ratio test. In the sequential analysis scenario, the test continues to operate as long as

$$B < l_n < A \tag{8}$$

Stop the test and decide for $\theta_0$ as soon as $l_n < B$, or decide for $\theta_1$ as soon as $l_n > A$. The constants $A$ and $B$ can be chosen by approximating $A \approx (1-\beta)/\alpha$ and $B \approx \beta/(1-\alpha)$ (Wald, 1947).

The randomized power martingale (3) is used to approximate the likelihood ratio (7), also a martingale. Consider the simple null hypothesis $H_0$: "no concept change in the data stream" against the alternative $H_1$: "concept change occurs in the data stream". Similar to (7), a ratio for a particular point $z_i$ is defined as

$$r(z_i) = \frac{f(z_i, H_1)}{f(z_i, H_0)} \approx \frac{\epsilon p_i^\epsilon}{p_i}$$

where the p-value $p_i$, by definition, is the smallest significance level (or observed size) at which a hypothesis will be rejected for a given observation. The p-value should not be interpreted as the probability that the null hypothesis $H_0$ is true (as a hypothesis is not a random event that can have a probability), despite a higher p-value when $H_0$ is true. We can, however, approximate $f(z_i, H_0)$ and $f(z_i, H_1)$ using $p_i$ and $\epsilon p_i^\epsilon$, respectively.

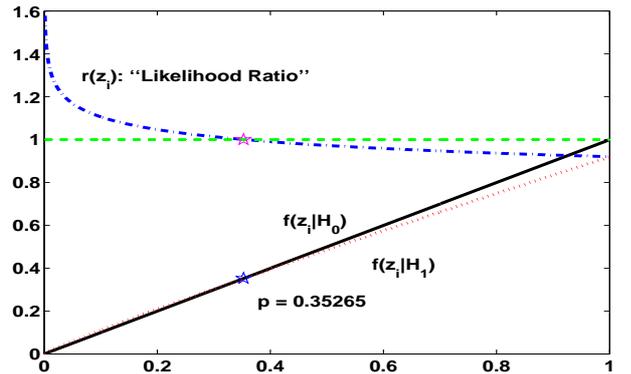

Figure 1: The "Likelihood Ratio" of a particular point $z_i$, derived from the p-values with $\epsilon = 0.92$ and when $\epsilon p_i^\epsilon - p_i = 0$, $p_i = 0.35265$. When $p_i < 0.35265$, $f(z_i, H_1) > f(z_i, H_0)$.

The characteristic of the approximated "likelihood ratio" at a particular point $r(z_i)$ is shown in Figure 1.

When $p_i < 0.35265$, the likelihood of a new concept is just slightly higher than the likelihood of an original concept. Hence, $r(z_i) > 1$ when $p_i < 0.35265$. The p-values output by (2) tend toward smaller values when concept change occurs (see Figure 1). As a result, $r(z_i)$ is likely to be greater than 1. A stream of $p_i < 0.35265$ will produce a stream of $r(z_i)$ greater than 1. The martingale $M_n^{(\epsilon)} = \prod_{i=1}^n r(z_i) \approx l_n$ will increase.

A higher $\epsilon$ shifts the intersection point of the approximated $f(z_i, H_1)$ and $f(z_i, H_0)$ to a higher (but $< 0.37$) p-value and increases the sensitivity of the martingale test. The test becomes more prone to false alarms. On the other hand, a lower $\epsilon$ increases the miss detection rate and lengthens the delay time.

By the Neyman-Pearson Fundamental Lemma, a decision in favor of the alternative hypothesis $H_1$, for some $\lambda \in \mathbf{R}^+$, happens when

$$f(Z, H_1) > \lambda f(Z, H_0) \qquad (9)$$

In our change-detection problem, we are only interested in the rejection of the null hypothesis. Hence, $B$ in (8) is set to 0 and $A$ is replaced by $\lambda$.

Following the argument in Wetherill and Glazebrook (1986), let $D$ be the set of all possible sets of $\{z_1, z_2, \cdots, z_n\}$ for all $n$, which lead to a decision in favor of the alternative hypothesis, then

$$\sum_{Z \in D} f(Z, H_1) = 1 - \beta \qquad (10)$$

$$\sum_{Z \in D} f(Z, H_0) = \alpha \qquad (11)$$

where $\alpha$ is the size of the test deciding in favor of the alternative hypothesis when the null hypothesis is true and $1 - \beta$ is the power of the test deciding in favor of the alternative hypothesis when it is true.

From (9), (10) and (11), we obtain the upper bound for $\lambda$,

$$\lambda \leq \frac{1-\beta}{\alpha} \qquad (12)$$

By equating $\lambda$ to this upper bound, our test is

$$\phi(z) = \begin{cases} \text{"change detected"}, & f(Z, H_1) > \frac{1-\beta}{\alpha} f(Z, H_0) \\ \text{"need more data"}, & f(Z, H_1) < \frac{1-\beta}{\alpha} f(Z, H_0) \end{cases}$$

Replacing the probability function by the martingale function with $Z$ implicit in the $p_i$ values,

$$\phi(z) = \begin{cases} \text{"change detected"}, & \prod_{i=1}^n \left(\epsilon p_i^{\epsilon-1}\right) > \frac{1-\beta}{\alpha} \\ \text{"need more data"}, & \prod_{i=1}^n \left(\epsilon p_i^{\epsilon-1}\right) < \frac{1-\beta}{\alpha} \end{cases}$$

At a fixed size $\alpha$, as the power of the test increases (i.e. $\beta$ decreases), the $\lambda$ value does not increase significantly. For example, at fixed $\alpha = 0.05$, $\lambda = 18$ when $\beta = 0.1$ and $\lim_{\beta \to 0} \lambda = \frac{1}{\alpha} = 20$. On the other hand, decreasing $\alpha$ at fixed $\beta$ increases $\lambda$ significantly. This observation suggests that the $\lambda$ value in the Doob's Maximal Inequality (6) is inversely proportional to the size $\alpha$ of the test and the power of the test is less prominently reflected in $\lambda$. A most powerful martingale test exists with $\lambda = \frac{1}{\alpha}$.

When $\alpha$ and $\beta$ are specified, the SPRT minimizes the average number of observations needed to make a decision when either of the hypothesis is true. In the case of the martingale test, we are interested in the mean (or median) time between the point change occurs and the point it is detected, known as the mean (or median) delay time, which can be estimated using the average sample number (ASN) (Wald, 1947) of a SPRT. When the alternative hypothesis $H_1$: "concept change occurs in the data stream" is true, the mean delay time, i.e. expected value of $n$ is:

$$E(n) = \frac{(1-\beta) \log \lambda}{E(\mathcal{L})} \qquad (13)$$

where

$$\mathcal{L} = \log \epsilon p_i^{\epsilon-1} \qquad (14)$$

Hence, with $\alpha$ and $\beta$ specified, the mean delay time depends on the expected value of $\mathcal{L}$, the logarithm of $r(z_i)$.

## 5 Experiments

Experiments are first performed on time-varying data streams simulated using two synthetic data sets to analyze the martingale test with respect to various $\lambda$ values in Section 5.1. We also show that the martingale test works well on high dimensional (i) numerical, (ii) categorical, and (iii) multi-class data streams simulated using three benchmark data sets in Section 5.2.

In the experiments, a fast adiabatic incremental SVM (Cauwenberghs & Poggio, 2000), using the Gaussian kernel and $C = 10$, is used to deduce the strangeness measure for the data points. A necessary condition for the test to work well is that the classifier must have a reasonable classification accuracy.

### 5.1 Simulated Data Streams using Synthetic Data Sets

In this subsection, we examine the performance of the martingale test based on the retrieval performance indicators, recall and precision, and the delay time for change detections for various $\lambda$ values on five different time-varying data stream scenarios simulated using the two synthetic data sets. The retrieval performance indicators, recall and precision, are defined in

our context as:

$$\text{Precision} = \frac{\text{Number of Correct Detections}}{\text{Number of Detections}}$$

Probability that a detection is actually correct, i.e. detecting a true change.

$$\text{Recall} = \frac{\text{Number of Correct Detections}}{\text{Number of True Changes}}$$

Probability that a change detection system recognizes a true change.

The delay time for a detected change is the number of time units from the true change point to the detected change point, if any.

First, we describe how the data streams with concept changes are simulated (i) using rotating hyperplane (Hulten et al., 2001) (Section 5.1.1), and (ii) using normally distributed clusters data generator (NDC) (Musicant, 1998) (Section 5.1.2) and then, the experimental results are presented in Section 5.1.3.

### 5.1.1 Simulated Data Stream using Rotating Hyperplane

Data stream is simulated using rotating hyperplane to generate a sequence of 100,000 data points consisting of changes occurring at points $(1,000 \times i) + 1$, for $i = 1, 2, \cdots, 99$. First we randomly generate 1,000 data points with each component values ranged in $[-1, 1]$. These data points are labeled positive and negative based on the following equation:

$$\sum_{i=1}^{m} w_i x_i = \begin{cases} < c & : \text{ negative} \\ \geq c & : \text{ positive} \end{cases} \quad (15)$$

where $c$ is an arbitrary fixed constant, $x_i$ is the component of a data point, $x$, and the fixed components, $w_i$, of a weight vector are randomly generated between -1 and 1. Similarly, the next 1,000 random data points are labeled using (15) with a new randomly generated fixed weight vector. This process continues until we get a data stream consisting of 100 segments, with each segment having 1,000 data points. Scenario (A), (B), (C) and (D) simulated using rotating hyperplane are summarized below:

- **Scenario (A) - Gradual Changes:** For $m = 2$, we control the rotation of the hyperplane about the origin by using the weights $w = [cos(r)\ sin(r)]$ restricting $r \in [-\frac{\pi}{3}, \frac{\pi}{3}]$. Hence, the maximum angle difference between two consecutive hyperplanes is $\frac{2\pi}{3}$ rad.

- **Scenario (B) - Arbitrary Changes (Gradual and Abrupt Changes):** For $m = 2$, we allow the weight $w = [cos(r)\ sin(r)]$ to change with $r \in [-\pi, \pi]$. In this case, any rotating hyperplane about the origin is possible.

- **Scenario (C) - Arbitrary Changes with Noisy Data Stream:** Noise is added to Scenario (B) by randomly switching the class labels of $p\%$ of the data points. In the experiment, $p = 5$.

- **Scenario (D) - Arbitrary Changes in a High Dimensional, Noisy Data Stream:** Setting $m = 10$, we repeat Scenario (C).

### 5.1.2 Simulated Data Streams using the Normally Distributed Clusters Data Generator (NDC)

Linearly non-separable binary-class data streams of 100,000 data points consisting of changes occurring at points $(1,000 \times i) + 1$, for $i = 1, 2, \cdots, 99$ is simulated using the NDC in $\mathbf{R}^{10}$ with randomly generated cluster means and variances. The values for each dimension are scaled to range in $[-1, 1]$. The generating process for the data stream is similar to that used for the rotating hyperplane data stream described in Section 5.1.1. Scenario (E) simulated using NDC is summarized below:

- **Scenario (E) : Arbitrary Changes in a High Dimensional, Noisy, Linearly Non-separable Data Stream:** Setting $m = 10$ and noise is added by randomly switching the class labels of $p\%$ of the data points. In the experiment, $p = 5$.

### 5.1.3 Results

Figure 2, 3, 4, 5 and 6 show the recall, precision and delay time of the martingale test on the five different data stream scenarios simulated using the rotating hyperplane and normally distributed clusters.

Some observations on the recall and precision of the martingale criteria for the different scenarios are reported below:

1. Precision decreases with decreasing $\lambda$ value. The precision shows similar trends, decreasing from 1.0 to around 0.8 as $\lambda^{-1}$ increases from 0.01 to 0.25, independent of the different scenarios.

2. The recall of a data stream with gradual changes (Scenario (A)) is lower than the recall of a data stream with arbitrary changes (Scenario (B)). The martingale test is more sensitive towards less restricted changes.

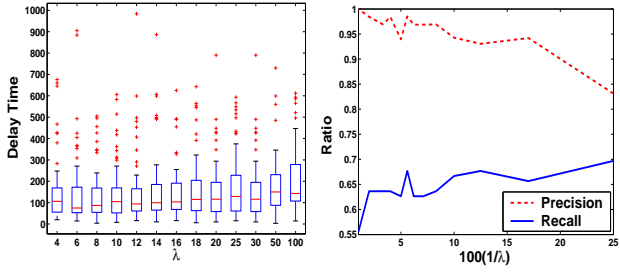

Figure 2: Scenario (A): Gradual changes.

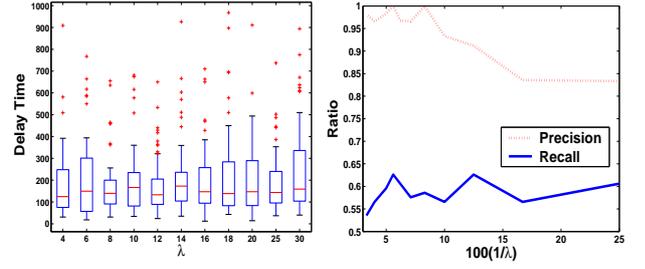

Figure 4: Scenario (C): Arbitrary changes with noise.

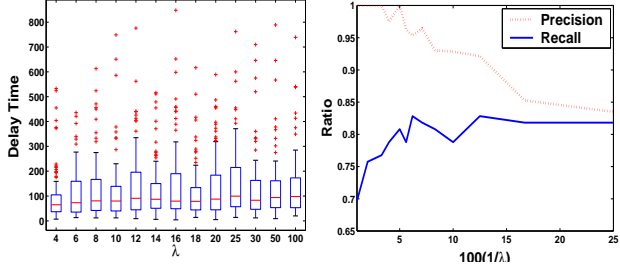

Figure 3: Scenario (B): Arbitrary changes (gradual and abrupt changes).

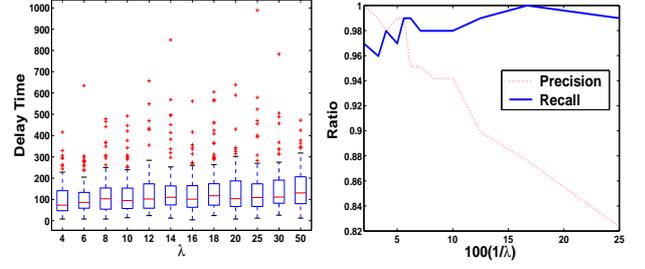

Figure 5: Scenario (D): Arbitrary changes on a high dimensional noisy data stream.

3. Both a noisy data stream (Scenario (C)) and a data stream with gradual changes (Scenario (A)) have low recall.

4. Higher dimension data streams (Scenario (D) and (E)) with arbitrary changes have high recall.

5. Despite Observation 3, noise in the high dimension data streams has limited effect on the recall.

As $\lambda$ increases from 4 to 100, the upper bound (6) becomes tighter, decreasing from 0.25 to 0.01. This corresponds to increasing precision, i.e. decreasing false alarm rate. All five scenarios reveal, unsurprisingly, that a higher precision, higher $\lambda$, comes at the expense of a higher median (or mean) delay time for the martingale test. As shown in the box-plots, the delay time distribution skews toward large values (i.e. small values are packed tightly together and large values stretch out and cover a wider range), independent of the $\lambda$. The delay time is very likely to be less than the mean delay time.

In real applications, $\lambda$ must be chosen to minimize losses (or cost) due to delay time, miss detections and false alarms.

T-tests are performed on the log transformed delay time of changes detected to determine if at a specific $\lambda$ value, say 10, two different scenarios could have the same (log) mean delay time. We observe that at more sensitive $\lambda$, i.e. lower $\lambda$ values, the means are likely to be equal. However, when noise is added, the mean delay time is affected. But if the data stream is high dimensional, noise has limited effects. For higher $\lambda$, say 30, the test is less sensitive but more cautious in detecting changes. The mean delay time varies for different scenarios. For Scenario (E), it is not surprising to see that its mean delay time is different from the other scenarios as the data set used is different. Hence, if the high dimensional data stream comes from the same source, a fixed $\lambda$ has similar performance in both a noiseless and a noisy environment. This may not be true for lower dimensional data stream.

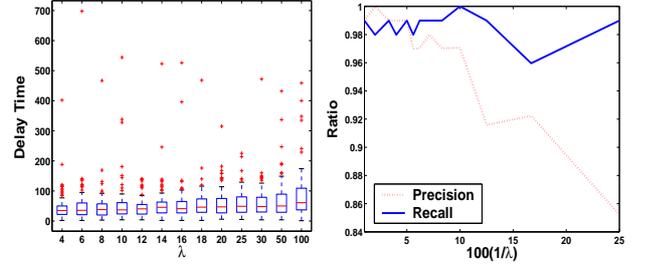

Figure 6: Scenario (E): Arbitrary changes on a high dimensional noisy linearly non-separable binary data stream.

### 5.2 Simulated Data Streams using Benchmark Data Sets

In this subsection, we describe how the data streams with concept changes are simulated using (i) combining ringnorm and twonorm data sets (Breiman, 1996) (Section 5.2.1), (ii) modifying UCI nursery data set (Blake & Merz, 1998) (Section 5.2.2), and (iii) modifying the USPS handwritten digits data set (LeCun et al., 1989) (Section 5.2.3). The effectiveness of the martingale test on high-dimensional (i) numerical, (ii) categorical, and (iii) multi-class data streams is shown in Figure 7, 8 and 9 for the simulated data streams using the above three benchmarks respectively.

### 5.2.1 Numerical High Dimensional Data sets: Ringnorm and Twonorm

We combined the *ringnorm (RN)* (two normal distribution, one within the other) and *twonorm (TN)* (two overlapping normal distribution) data sets to form a new binary-class data stream of 20 numerical attributes consisting of 14,800 data points. The 7,400 data points from RN is partitioned into 8 subsets with the first 7 subsets ($RN_i, i = 1, \cdots, 7$) consisting of 1,000 data points each and $RN_8$ consists of 400 data points. Similarly, the 7,400 data points from TN is also partitioned into 8 subsets with the first 7 subsets ($TN_i, i = 1, \cdots, 7$) consisting of 1,000 data points each and $TN_8$ consists of 400 data points.

The new data stream is a sequence of data points arranged such that $RN_1 TN_1 \cdots RN_7 TN_7 RN_8 TN_8$. This is done to simulate 15 changes in the structure of the data sequence at the points, $1000i + 1$, for $i = 1, \cdots, 14$ and the last one at 14,401.

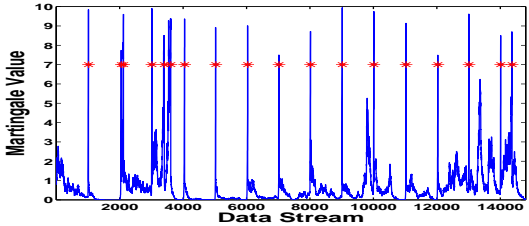

Figure 7: Simulated data stream using Ringnorm and Wavenorm: The martingale values of the data stream using $\lambda = 10$. ∗ represents detected change point: 1020, 2054, 2125 (false alarm) 3017, 3410 (false alarm), 3614 (false alarm), 4051, 5030, 6036, 7023, 8018, 9019, 10014, 11031, 12032, 13013, 14014, 14374 (false alarm) and one miss detection.

### 5.2.2 Categorical High Dimensional Data set: Nursery benchmark

We modified the nursery data set, consisting of 12,960 data points in 5 classes with 8 nominal attributes, to form a new binary-class data stream. First, we combined three classes (not recommended, recommended and highly recommended) into a single class consisting of 4,650 data points labeled as negative examples. The set $RN$ is formed by randomly selecting 4,000 out of the 4,650 data points. The "priority" class contains 4,266 data points are labeled as positive examples and we randomly select 4,000 out of the 4,266 data points to form the set $PP$. The "special priority" class, containing 4,044 data points are split into two subsets consisting of 2,000 data points each, a set ($SPP$) of positive examples and a set ($SPN$) of negative examples. The other 44 data points are removed.

New subsets of data points are constructed as follows:

- Set $A_i$: 500 negative examples from $RN$ and 500 positive examples from $PP$.
- Set $B_i$: 500 negative examples from $SPN$ and 500 positive examples from $PP$.
- Set $C_i$: 500 negative examples from $RN$ and 500 positive examples from $SPP$.

The data stream $S$ is constructed as follow:

$$S = A_1 B_1 C_1 A_2 B_2 C_2 A_3 B_3 C_3 A_4 B_4 C_4$$

consisting of 12,000 data points with 11 change points and in each subset, the data point positions are randomly permutated.

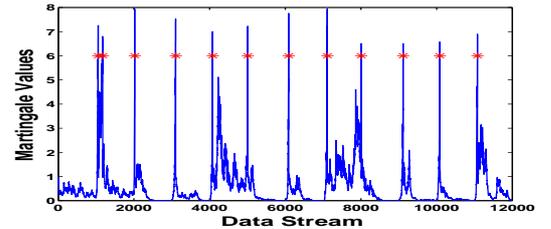

Figure 8: Simulated data stream using a modified Nursery data set: The martingale values of the data stream with $\lambda = 8$. ∗ represents detected change point: 1059, 1179 (false alarm), 2024, 3109, 4084, 5010, 6090, 7114, 8007, 9115, 10086, 11079

### 5.2.3 Multi-class High Dimensional Data: Three-digit data stream from USPS handwritten digits data set.

The USPS handwritten digits data set, consisting 7,291 data points in 10 classes with 256 numeric attributes, is modified to form a data stream as follows. There are four different data segments. Each segment draws from a fixed set of three different digits in a random fashion. The three-digit sets change from one segment to the next. The composition of the data stream and ground truth for the change points are summarized in Table 2. We note that the change points do not appear at fixed intervals. The one-against-the-rest multi-class SVM is used to extract p-values.

For the three-digit data stream, three one-against-the-rest SVM are used. Hence, three martingale values are computed at each point to detect change (see Figure 9). When one of the martingale values is greater than $\lambda$, change is detected.

| Segment | Digit 1 | Digit 2 | Digit 3 | Total | Change Point |
|---|---|---|---|---|---|
| 1 | 597 (0) | 502 (1) | 731 (2) | 1830 | 1831 |
| 2 | 597 (0) | 658 (3) | 652 (4) | 1907 | 3738 |
| 3 | 503 (1) | 556 (5) | 664 (6) | 1723 | 5461 |
| 4 | 645 (7) | 542 (8) | 644 (9) | 1831 | - |

Table 2: **Three-Digit Data Stream**: The number of data points are shown with the true digit class of the data points inside ( ).

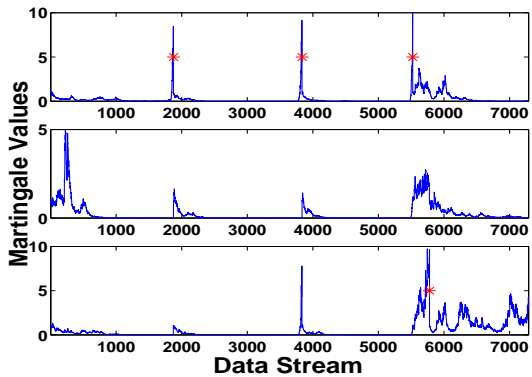

Figure 9: Simulated three-digit data stream using the USPS handwritten digit data set: The martingale values of the data stream. ∗ represents detected change point: 1876, 3837, 5523, 5780 (false alarm). True change points are 1831, 3738 and 5461. Delay time are 45, 99 and 62.

## 6 Conclusions

In this paper, we describe the detection of concept change in time-varying data streams by testing whether exchangeability condition is violated using martingale. The contribution of the paper is the justification that this change-detection martingale test based on the Doob's Maximal Inequality is an approximation of the sequential probability ratio test (SPRT), which is used to (i) deduce the relation between the threshold value used in our change-detection test and its size and power, and (ii) estimate the mean delay time before a change is detected.

Future works include (i) building a robust adaptive learning system based on the martingale test, and (ii) extending the martingale test to detect changes in unlabeled data streams and single-class data streams.

**Acknowledgments**

The first author thanks the reviewers for useful comments, and Alex Gammerman and Vladimir Vovk for the manuscript of (Vovk et al., 2005), and useful discussions.